\documentclass[10pt,twocolumn,letterpaper]{article}

\usepackage{cvpr}    

\usepackage{graphicx}
\usepackage{amsmath}
\usepackage{amssymb}
\usepackage{booktabs}
\usepackage{lipsum}
\usepackage{float}  %

\usepackage[accsupp]{axessibility}  %

\usepackage{pifont}
\newcommand{\cmark}{\ding{51}}%
\newcommand{\xmark}{\ding{55}}%
\newcommand{\rev}[1]{\textcolor{black}{#1}}

\usepackage{xcolor}

\usepackage[pagebackref,breaklinks,colorlinks]{hyperref}

\usepackage{color}

\usepackage[capitalize]{cleveref}
\crefname{section}{Sec.}{Secs.}
\Crefname{section}{Section}{Sections}
\Crefname{table}{Table}{Tables}
\crefname{table}{Tab.}{Tabs.}

\def\cvprPaperID{1088} %
\def\confName{CVPR}
\def\confYear{2022}

\definecolor{somegray}{rgb}{0.5, 0.5, 0.5}
\newcommand{\darkgrayed}[1]{\textcolor{somegray}{#1}}
\makeatletter
\newcommand*\titleheader[1]{\gdef\@titleheader{#1}}
\AtBeginDocument{%
  \let\st@red@title\@title
  \def\@title{%
    \vskip-3em
    \bgroup\normalfont\large\centering\@titleheader\par\egroup
    \vskip1.5em\st@red@title}
}

\makeatother

\titleheader{\darkgrayed{This paper is an updated version of the paper published at the
IEEE Conference\\ on Computer Vision and Pattern Recognition (CVPR), New Orleans, 2022. \copyright IEEE \\
It features updated computational complexity calculations in Sec. 5.}}

\makeatletter
\let\@oldmaketitle\@maketitle%
\renewcommand{\@maketitle}{\@oldmaketitle
\centering
\includegraphics[width=1.0\textwidth]{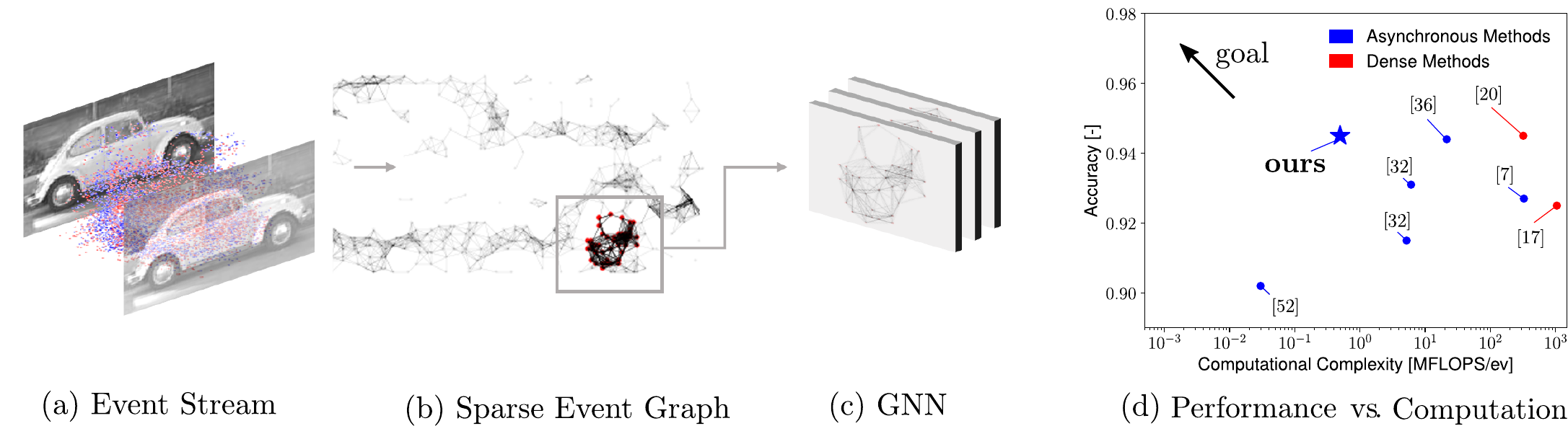}
\vspace{-1ex}
\captionof{figure}{Our method processes events (a) as spatio-temporal graphs (b), using a graph neural network (c). For each new event we only limit computation to a local subgraph of (b), corresponding to the receptive field of the network, thus significantly reducing per-event computation. Our method has a 11 times lower computational complexity than state-of-the-art asynchronous method\cite{Li21iccv}, while achieving state-of-the-art results on object recognition and object detection (d). 
\vspace{0.3cm}}
\label{fig:eyecatcher}
}%
\makeatother

\title{AEGNN: Asynchronous Event-based Graph Neural Networks}

\begin{document}

\author{Simon Schaefer\thanks{these authors contributed equally}
\quad\quad\quad
Daniel Gehrig$^*$
\quad\quad\quad
Davide Scaramuzza \\ \\
Dept. Informatics, Univ. of Zurich and \\
Dept. of Neuroinformatics, Univ. of Zurich and ETH Zurich
}
\maketitle

\begin{abstract}
   The best performing learning algorithms devised for event cameras work by first converting events into dense representations that are then processed using standard CNNs. However, these steps discard both the sparsity and high temporal resolution of events, leading to high computational burden and latency. For this reason, recent works have adopted Graph Neural Networks (GNNs), which process events as ``static" spatio-temporal graphs, which are inherently "sparse". We take this trend one step further by introducing Asynchronous, Event-based Graph Neural Networks (AEGNNs), a novel event-processing paradigm that generalizes standard GNNs to process events as ``evolving" spatio-temporal graphs. 
      AEGNNs follow efficient update rules that restrict recomputation of network activations only to the nodes affected by each new event, thereby significantly reducing both computation and latency for event-by-event processing. 
      AEGNNs are easily trained on synchronous inputs and can be converted to efficient, "asynchronous" networks at test time. We thoroughly validate our method on object classification and detection tasks, where we show an up to a 11-fold reduction in computational complexity (FLOPs), with similar or even better performance than state-of-the-art asynchronous methods. This reduction in computation directly translates to an 8-fold reduction in computational latency when compared to standard GNNs, which opens the door to low-latency event-based processing. 
\end{abstract}

\section*{Multimedia Material}
For videos, code and more, visit our project page \url{https://uzh-rpg.github.io/aegnn/}.

\section{Introduction}
\label{sec:intro}
Compared to standard frame-based cameras, which measure absolute intensity at a synchronous rate, event-cameras only measure \emph{changes} in intensity, and do this independently for each pixel, resulting in an asynchronous and binary stream of \emph{events} (Figure ~\ref{fig:eyecatcher} (a)). 
These events measure a highly compressed representation of the visual signal and are characterized by microsecond-level latency and temporal resolution, a high dynamic range of up to 140 dB, low motion blur, and low power (milliwatts instead of watts).
Due to these outstanding properties, event cameras are indispensable sensors in challenging application domains---such as robotics\cite{Falanga20science,Sanket20icra,Sun21ral,Hagenaars20ral}, autonomous driving\cite{Gehrig21ral,Zhu18ral,Sironi18cvpr}, and computational photography\cite{Rebecq19pami,Tulyakov21cvpr,Bardow16cvpr,Zhang21cvpr}---characterized by frequent high-speed motions, low-light and high-dynamic-range scenes, or in always-on applications, where low power is needed, such as IoT video surveillance\cite{Mitra09tbcs,Indiveri11fns}. A survey about applications and research in event-based vision can be found in\cite{Gallego20pami}.
    
The output of event cameras is inherently sparse and asynchronous, making them incompatible with traditional computer-vision algorithms designed for standard images. This prompts the development of novel algorithms that optimally leverage the sparse and asynchronous nature of events. 
In doing so, existing algorithms designed for event cameras have traded off latency and prediction performance.
\emph{Filtering-based} \cite{Orchard15pami} \cite{Lagorce17pami} approaches process events sequentially, and, thus, can provide low-latency predictions and a high temporal resolution. 
However, these approaches usually rely on handcrafted filter equations, which do not scale to more complex tasks, such as object detection or classification.
Spiking Neural Networks (SNNs) are one instance of filtering-based models, which seek to learn these rules in a data-driven fashion, but are still in their infancy, lacking general and robust learning rules \cite{Gehrig20icra,Shrestha18nips,Lee16fns}. As a result, SNNs typically fail to solve more complex high-level tasks\cite{Orchard15pami,Sironi18cvpr,PerezCarrasco13pami,Amir17cvpr}. 
Many of the challenges above can be avoided by processing events as batches. In fact, 
recent progress has been made by converting batches of events into \emph{dense}, image-like representations and processing them using methods designed for images, such as convolutional neural networks (CNNs).
By adopting this paradigm, learning-based methods using CNNs have made significant strides in solving computer vision tasks with events \cite{Hidalgo20threedv,Maqueda18cvpr,Tulyakov21cvpr,Rebecq19pami,Gehrig19iccv,Zhu18rss,Zhu19cvpr,Perot20nips}.

However, while easy to process, treating events as image-like representations discards their sparse and asynchronous nature and leads to wasteful computation. This wasteful computation directly translates to higher power consumption and latency\cite{Indiveri11fns,Moradi18tbcs,Aimar19tnnls}. A recent line of work\cite{Chang20jestcs} showed on an FPGA that by reducing the computational complexity by a factor of 5, they could reduce the latency by a factor of 5 while reducing the power consumption by a factor of 4. Therefore, by eliminating wasteful computation, we can expect significant decreases in the power consumption and latency of learning systems.

Currently, this wasteful computation is caused by two factors: On the one hand, due to the working principle of event cameras, they trigger predominantly at edges, while large texture-less or static regions remain without events. Image representations typically encode these regions as zeros, which are then unnecessarily processed by standard neural networks.  
On the other hand, for each new event, standard methods would need to recompute all network activations. However, events only measure single pixel changes and, thus, leave most of the activations unchanged, leading to unnecessary recomputation of activations.

A recent line of work seeks to address both of these challenges by reducing the computational complexity of learning-based approaches while maintaining the high temporal resolution of events. 
A key ingredient to keeping high performance in this setting was the adoption of geometric learning methods, such as recursive point-cloud processing\cite{Sekikawa19cvpr} or Asynchronous Sparse Convolutions\cite{Messikommer20eccv}.  In both works, standard neural networks 
were trained using batches of events, leveraging well-established learning techniques such as backpropagation, and then deploying them in an \emph{event-by-event} fashion at test time, thus minimizing computation.
However, both of these methods suffer from limitations: While \cite{Sekikawa19cvpr} does not perform hierarchical learning, limiting scalability to complex tasks, \cite{Messikommer20eccv}, relies on a specific type of input representation, which discards the temporal information of events. 

In this work, we introduce Asynchronous, Event-based Graph Neural Networks (AEGNN), a neural network architecture geared toward processing events as graphs in a sequential manner (Fig. \ref{fig:eyecatcher}).
For each new event, our method only performs local changes to the activations of the GNN, and propagates these asynchronously to lower layers. Similar to \cite{Messikommer20eccv,Sekikawa19cvpr}, AEGNNs can be trained on batches of events---thus leveraging backpropagation---and can later be deployed in an asynchronous mode, generating the identical output. However, they address the key limitations of previous work: \emph{(i)} They allow hierarchical learning using standard graph neural networks and \emph{(ii)} model events as spatio-temporal graphs, thus retaining their temporal information, instead of discarding it. This leads to significant computational savings.
We summarize our contributions as follows:
\begin{itemize}
    \item We introduce AEGNN, a novel paradigm for processing events sparsely and asynchronously as temporally evolving graphs. This allows us to process events efficiently, without sacrificing their sparsity and high temporal resolution. 
    \item (ii) We derive efficient update rules, which allow us to simply train AEGNNs on synchronous event-data, and then deploy them in an asynchronous mode during test-time. These rules are general and can be applied to most existing graph neural network architectures.
    \item (iii) We apply AEGNNs on object recognition and object detection benchmarks. For object detection, we show similar performance to state-of-the-art methods, while requiring up to 11 times less compute, while for object detection we show a 32\% computation reduction with an up to 3.4\% increase in terms of mAP.
\end{itemize}

\begin{figure*}[t]
\centering
\begin{tabular}{c}
\includegraphics[width=\textwidth]{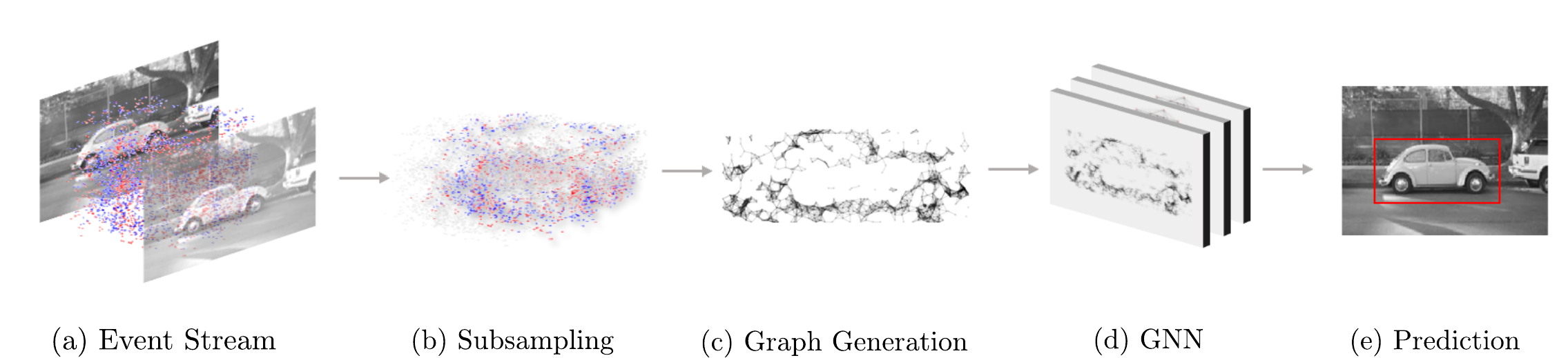}\\
\includegraphics[width=\textwidth]{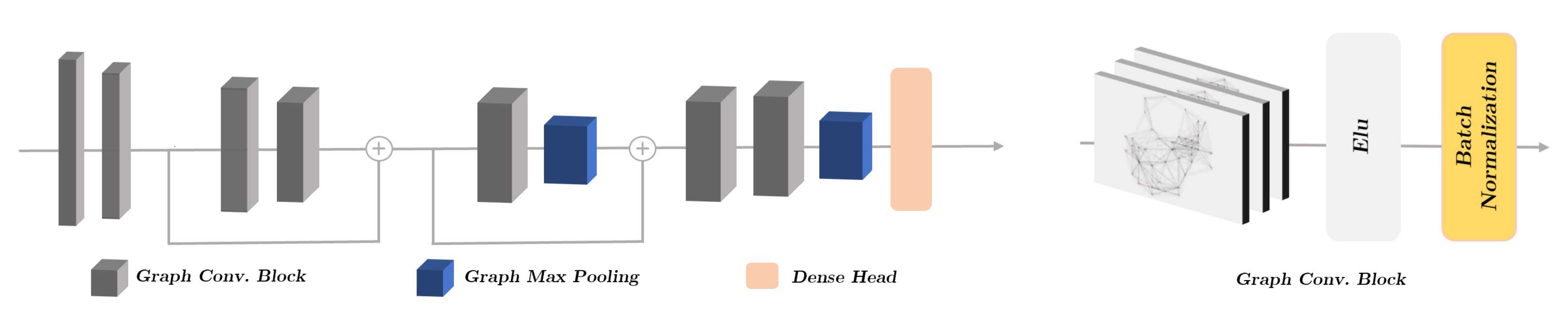}
\end{tabular}
\caption{Overview of the processing steps in our method. The event stream (a) is first subsampled using uniform sampling (b). The subsampled events are used to generate a sparse spatio-temporal graph (c), which is processed by a graph neural network (GNN)(d), which generates a bounding-box prediction (d). 
Although our method works for any task, here we illustrate our method for the task of object detection. 
In the figure below, we show an overview of the used network architecture. It combines Graph Convolutions (here Spline Convolutions) with pooling layers, followed by a prediction head.  Each graph convolution block consists of several graph convolutions followed by ELU and Batch Normalization.}
\label{fig:pipeline}
\end{figure*}

\section{Related Work}
Since the advent of deep learning, event-based vision has adopted many of its models. Early models, relied on shallow learning techniques such as SVMs\cite{Sironi18cvpr} or filtering-based techniques\cite{Orchard15pami,Lagorce17pami,Gallego17pami,Kim16eccv}, and have gradually shifted to deeper architectures such as CNN's\cite{Maqueda18cvpr,Gehrig19iccv,Zhu18rss,Rebecq19pami}. 
While achieving state-of-the-art performance, these types of models do not take into account the sparse and asynchronous nature of events, leading to redundant computation. 
This prompted the development of sparse network architectures such as SNNs, point cloud methods\cite{Sekikawa19cvpr}, Submanifold Sparse Convolutions\cite{Messikommer20eccv} and graph neural networks\cite{Bi19iccv,Bi20trip,Li21iccv}. Which all seek to reduce computation. 
While SNNs are traditionally harder to train, due to a lack of efficient learning rules, geometric learning methods such as \cite{Sekikawa19cvpr,Bi19iccv,Bi20trip,Li21iccv,Messikommer20eccv} have gained popularity in recent years, since they are more suited to the asynchronous and sparse nature of events, and are easily trained and implemented thanks to the existence of well-maintained toolboxes.

In particular, graph-based methods such as \cite{Bi19iccv,Bi20trip,Li21iccv,Rossi20icmlw} show a significant reduction in computational complexity compared to dense methods that rely on standard CNNs. This is because, instead of processing events as dense image-like tensors, they only consider sparse connections between events, and confine message passing to these connections.  
Despite this sparsity, these methods still process events as batches and thus need to recompute all activations, whenever a new event arrives. However, each event only indicates a per-pixel change, and thus recomputing activations leads to the highly redundant computation. 
To counteract this, a recent line of work has focused on reusing network activations as much as possible between consecutive events, by applying efficient recursive update rules\cite{Sekikawa19cvpr} and propagating these to lower layers\cite{Messikommer20eccv}. 

These methods, however, do not allow for hierarchical learning\cite{Sekikawa19cvpr, Rossi20icmlw} or still rely on sparse but image-like input representations, which discard the temporal component of events. 
These factors either limit the scalability to more complex tasks in the case of \cite{Sekikawa19cvpr}, or degrade performance while incurring higher computation in the case of \cite{Messikommer20eccv}.
\rev{Most similar to our work, \cite{Rossi20icmlw} learns on dynamic graphs, by performing learned updates each time node events are triggered. However, it also performs shallow learning, \emph{i.e.} it only computes node embeddings, but does not use them for end-task learning.}

In this work, we combine the advantages of graph-based methods with efficient recursive update rules, thus addressing these limitations: Asynchronous Event-based Graph Neural Networks are multi-layered, and can thus learn more complex tasks than\cite{Sekikawa19cvpr}, and leverage the spatio-temporal sparsity of events better than \cite{Messikommer20eccv}, leading to significant computation reduction.

\section{Prerequisites}
\label{sec:prerequisites}
In this work, we model events as spatio-temporal graphs $\mathcal{G} = \{\mathcal{V}, \mathcal{E}\}$ with vertices $\mathcal{V}$ and (directed) edges $\mathcal{E}$. In this context, events are represented as nodes within the graph and connections are formed between neighboring events (Fig.\ref{fig:pipeline} (c)). 
We use a graph neural network to process this graph and generate a prediction $y$. It can be represented as a function $f(\mathcal{G}) =y$, which executes a set of operations on the graph level. 
Most common operations consist of graph convolutions and pooling steps, which operate on node features $\mathbf{x}_i$ attached to each node, and edge features $e_{ij}$ attached to each edge.

\textbf{Graph Convolutions:} Graph convolutions generally consist of three distinct steps which are repeated for each node $i$ in the graph: First the function $\psi$ computes messages based on pairs of neighbors $(i,j)$, where $i$ is fixed and $j\in\mathcal{N}(i)$ is in the neighborhood of $i$. These messages depend on the node features at these nodes, the edge feature but also on the spatial arrangement of nodes $i$ and $j$. Next, all messages are aggregated through summation\footnote{While summation is the most common form of aggregation, any function which is symmetric in its inputs can be used, such as $\max$ and $\min$ or $\sum$.}, and followed by a function $\gamma_{\Theta}$, which computes the new value for node $i$. 
These steps are summarized in the equations below:
\begin{align}
    \mathbf{z}_i&=\sum_{j \in \mathbb{N}(i)} \psi_\Theta (\mathbf{x}_i, \mathbf{x}, \mathbf{e}_{ij})\\
    \hat{\mathbf{x}}_i &= \gamma_\Theta (\mathbf{x}_i, \mathbf{z}_i)
    \label{eq:graph_conv}
\end{align}
Both $\psi$ and $\gamma$ denote differentiable functions such as a multi-layer perceptron, parametrized by $\Theta=\{\theta_\gamma, \theta_\psi\}$.

\textbf{Graph Pooling} Graph pooling operations transform a graph $\mathbb{G}$ to a more coarse graph $\mathbb{G}_c$. For an overview of the different types of graph pooling, we refer to \cite{Zhou20aiopen}. Within this work, we will focus on cluster-based pooling methods, which aggregate the graph nodes into clusters $\mathcal{C}_k$ with cluster centers $k\in\mathcal{V}_c$ which form a subset of $\mathcal{V}$. The new features at these cluster centers are computed by aggregating features in each cluster:  
\begin{align}
    \mathbf{x}_k = \max_{i \in \mathcal{C}_k} \mathbf{x}_i 
    \label{eq:graph_pooling}
\end{align}
Since clustering reduces the number of nodes, the original edges need to reconnected, and this is performed with the function $\pi$:
\begin{align}
    \mathbb{E}_{c} = \pi(\mathbb{E}, \mathbb{C})
    \label{eq:graph_edge_pooling}
\end{align}
resulting in the final coarse graph. \\
Stacking these operations as layers enables rich, and high-level feature computation, making these models more powerful than the point cloud method in \cite{Sekikawa19cvpr} or shallow features computed in \cite{Sironi18cvpr}.

\section{Approach}
\label{sec:approach}
Representing event data as spatio-temporal graphs allows us to efficiently process incoming events by performing sparse but complete graph updates. In the following, we show how a graph can be constructed from an event stream (Sec. \ref{sec:method/graph_construction}), and we demonstrate how it can be used for efficient and asynchronous computations (Sec. \ref{sec:method/asy_processing}). An overview of the full method is illustrated in Figure \ref{fig:pipeline}.

\subsection{Graph Construction}
\label{sec:method/graph_construction}
Event cameras have independent pixels which each trigger events, whenever they perceive a brightness change. Each event encodes the pixel position $(x_i,y_i)$, time $t_i$ with microsecond level resolution and polarity (sign) $p_i\in\{-1,1\}$ of the change. A group of event in a time window $\Delta T$, can thus be represented as an ordered list of tuples 
\begin{align}
    \{e_i\}_N = \{e_i\}_{i=1}^N\quad\text{ with }e_i=(x_i, y_i, t_i, p_i)  
\end{align}
By embedding these events in a spatio-temporal space $\mathbb{R}^3$ we thus can see that they are inherently sparse and asynchronous (Fig.\ref{fig:pipeline} (a,b)). \\
For the sake of computational efficiency, we first sub-sample the events uniformly by a factor $K$ (Fig. \ref{fig:pipeline} (b)). In this work, we select $K=10$. \rev{While this preprocessing step removes events, we found that it is critical to combat overfitting, since the network learns to consider larger contexts, focusing on more informative events. }In contrast to other representations of event data such as event histograms \cite{Messikommer20eccv} or event volumes \cite{Perot20nips, Bardow16cvpr}, the full temporal resolution of the event stream is preserved. This high temporal resolution is crucial in robotic applications like obstacle avoidance\cite{Sanket20icra,Falanga20science,Loquercio2021Science}.

We use the remaining events to form an event graph $\mathcal{G}$, where each event is a node (Fig. \ref{fig:pipeline} (c)). Inspired by \cite{Bi19iccv} the event's temporal position is normalized by a factor $\beta$ to map it to a similar range as the spatial coordinates. The position of each vertex is then denoted as $\mathbf{X}_i=(x_i, y_i, t_i^*)$ with $t^*_i = \beta t_i$. \\
For each pair of nodes $i$ and $j$, an edge $e_{ij}$ between them is generated if they are within spatio-temporal distance $R$, \emph{i.e.} $R\geq \Vert\mathbf{X}_i-\mathbf{X}_j\Vert$ from each other. To reduce computation and regularize the graph, we limit the maximal number of neighborhood nodes to $D_{max}$, i.e. $|\mathcal{N}(i)| \leq D_{max}$.
Finally, we assign initial node features, $\mathbf{x}_i=p_i$ and edge features corresponding to the relative position between the connected vertices, normalized by $R$.

\subsection{Asynchronous Processing}
\label{sec:method/asy_processing}
As we slide the time window $\Delta T$, new events enter this window, and old events leave the window. While traditional methods would need to recompute all activations once this happens, here we present a recursive formulation that incorporates new events with minimal computation.

\begin{figure}[H]
\centering
\includegraphics[width=\linewidth]{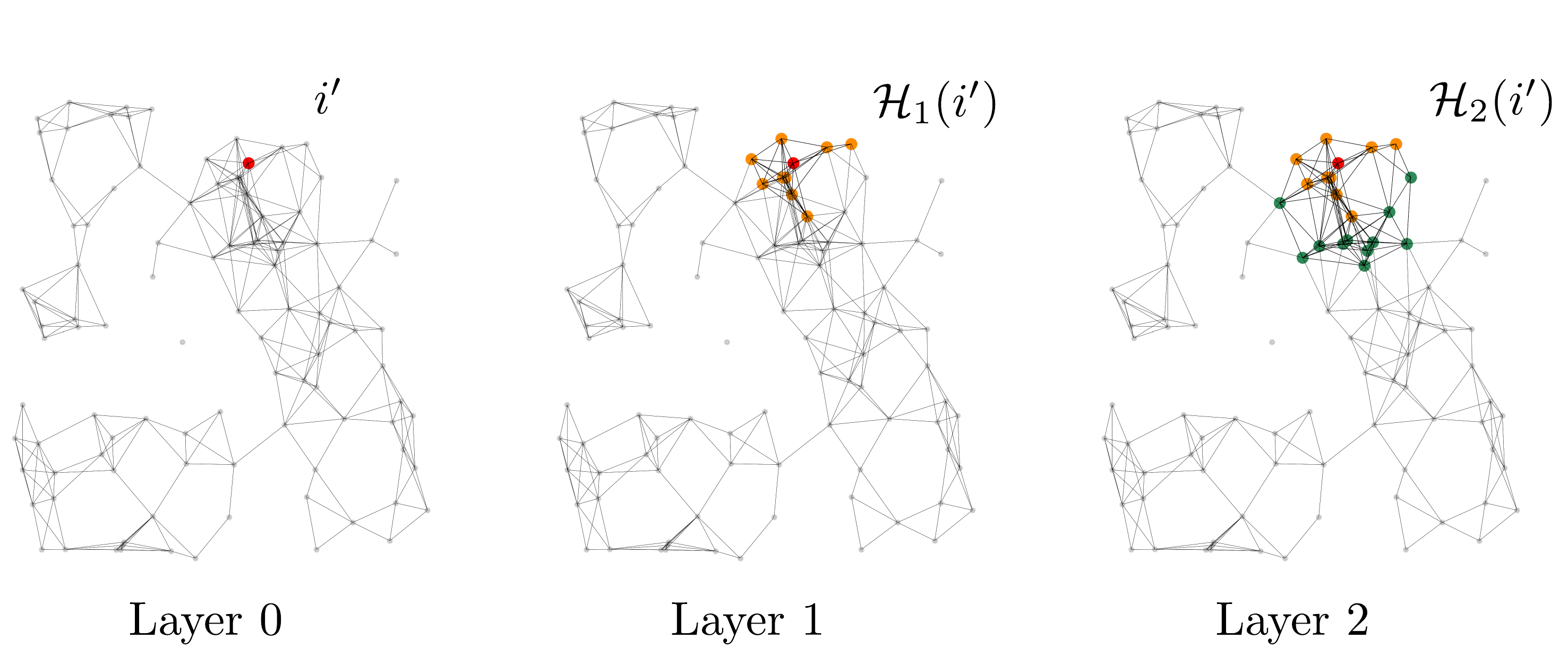}
\caption{Message propagation in the event graph. A new event (red) is generated and added to the graph of precedent events (left). The added information is propagated to the $k$-hop neighborhood of the new event vertex, with $k = 1$ (middle) and $k = 2$ (right).}
\label{fig:graph_neighbors}
\end{figure}

As a new event arrives, a new node is added to the graph, together with new edges connecting this node to existing vertices. The new connections are sparse, affecting only neighboring events. In fact, in the first layer, a new event only affects the state of its 1-hop subgraph (Fig.\ref{fig:graph_neighbors}, Layer 1), corresponding with the neighborhood of the new node $i'$. 
Therefore, activations in the next layer need to only be recomputed for this subgraph via Eq. \eqref{eq:graph_conv}. 
\begin{align}
    \hat{\textbf{z}}_i &= \sum_{j\in\mathcal{N}(i)} \psi_\Theta(\textbf{x}_i,\textbf{x}_j,\textbf{e}_{ij})\\
    \hat{\textbf{x}}_i&=\gamma_\Theta(\textbf{x}_i,\textbf{z}_i) \text{ for all } i \in \mathcal{N}(i')
\end{align}
As deeper layers are reached, this subgraph expands, hopping one node after each layer step, until at layer $N$ the nodes in $\mathcal{H}_N(i')$ need to be updated. 
$\mathcal{H}_N(i')$ denotes the N-hop subgraph which contains all nodes $j$ such that $j$ could be reached from $i'$ using $N$ hops or fewer. 
We visualize this hopping behavior in Fig.~\ref{fig:graph_neighbors}. 
Instead of processing the whole graph, only this subgraph has to be processed to obtain the same resulting graph activations as Eq. \ref{eq:graph_conv}. 
By iteratively applying this concept to each graph-convolution layer of a graph neural network, its forward pass can be formulated sparsely, which significantly reduces the computational effort.
At each layer, the necessary computation is proportional to the number of nodes in the respective subgraph. This number is known in the graph-theory literature as \emph{neighborhood function}\cite{Boldi11ACM}, and is influenced by the average and variance of the connectivity of the graph, which together forms the \emph{index of dispersion}\cite{Boldi11ACM}.

\begin{table*}[t]
    \centering
    \begin{tabular}{lccccccc}
    & & & \multicolumn{2}{c}{N-Caltech101} & & \multicolumn{2}{c}{N-Cars} \\
    \cline{4-5} \cline{7-8}
    Methods & Representation & Async. & Accuracy $\uparrow$ & MFLOP/ev $\downarrow$ & &  Accuracy $\uparrow$ & MFLOP/ev $\downarrow$ \\
    \hline
    H-First \cite{Orchard15pami} & Spike & \cmark & 0.054 & - & & 0.561 & - \\
    HOTS \cite{Lagorce17pami} & Time-Surface & \cmark & 0.210 & 54.0 & & 0.624 & 14.0 \\
    HATS \cite{Sironi18cvpr} & Time-Surface & \cmark & 0.642 & 4.3 & & 0.902 & 0.03 \\
    DART \cite{Ramesh17arxiv} & Time-Surface & \cmark & 0.664 & - & & - & - \\
    YOLE \cite{Cannici19cvprw} & Event-Histogram & \cmark & 0.702 & 3659 & & 0.927 & 328.16 \\
    EST \cite{Gehrig19iccv} & Event-Histogram &\xmark & \textbf{0.817} & 4150 & & 0.925 & 1050 \\
    SSC \cite{Graham18cvpr} & Event-Histogram &\xmark & 0.761 & 1621 & & 0.945 & 321 \\
    AsyNet \cite{Messikommer20eccv} & Event-Histogram & \cmark & 0.745 & 202 & & 0.944 & 21.5 \\
    NVS-S \cite{Li21iccv} & Graph & \cmark & 0.670 & 7.8 & & 0.915 & 5.2 \\
    EvS-S \cite{Li21iccv} & Graph & \cmark & 0.761 & 11.5 & & 0.931 & 6.1 \\
    \hline
    \textbf{Ours} & Graph & \cmark & 0.668 & \textbf{7.31} & & \textbf{0.945} & \textbf{0.47}
    \end{tabular}
    \caption{Comparison with several asynchronous and dense methods for object recognition. Our graph-based method has the lowest computational complexity overall while achieving state-of-the-art performance. Especially, it obtains the best accuracy on N-Cars\cite{Sironi18cvpr} with $20$ times lower computational complexity, compared to the second-best asynchronous method.}
    \label{tab:performance_recognition}
\end{table*}

\textbf{Graph Convolutions} Our sparse update rules for graph convolutions are agnostic to the choice of functions $\psi$ and $\gamma$ (Eq. \ref{eq:graph_conv}) and are therefore applicable to arbitrary types of graph convolution. It consists of two steps: During the \textit{initialization} the convolution is applied to the full graph, while the resulting graph, i.e., the vertices and edges as well as their attributes, are stored. \rev{We perform this step at the beginning and whenever the camera is stationary and mostly noise events enter the sliding window.} Thereafter, in the \textit{processing} step, every time a new vertex is inserted into the graph, the graph only changes locally. Therefore, a full graph update is equivalent to updating the 1-hop subgraph starting from the new vertex, by applying Eq.  \ref{eq:graph_conv} to its 1-hop subgraph only. Thereby, the subgraph can be efficiently obtained, as the graph's edges are known from the initialization and updated with every subsequent forward pass. \\
The same procedure can be applied to every subsequent convolutional layer. Hence, the update of the kth layer is limited to the $k$-hop subgraph of the new vertex. These steps lead to significant computational savings, as demonstrated in Sec. \ref{sec:experiments}.

\textbf{Graph Pooling} Similar to sparse graph convolutions, sparse graph pooling operations are composed of an \textit{initialization} and a \textit{processing} step. During \textit{initialization}, the procedure described in Sec. \ref{sec:prerequisites} is applied to the dense input graph $\mathcal{G}$, which results in the coarse output graph $\mathcal{G}_c$. Subsequently, in the \textit{processing} stage, \rev{we assign events to the respective voxels where they are triggered, connecting them with nodes in the input graph, and then perform the max operation again for that specific voxel. If a node attribute is changed, we similarly perform the max operation again at the respective voxel.} Finally, the output graph $\mathcal{G}_c$ can be efficiently computed by applying Eqs. \ref{eq:graph_pooling} and \ref{eq:graph_edge_pooling} on $\mathcal{G}_c'$. 

\textbf{Other Layers} Non-graph-based layers such as linear or batch normalization can be sparsely updated similarly, by storing the results of the dense update during initialization and only processing the subset of the input, which changes from the previous input, as described in \cite{Messikommer20eccv}. However, since these layers are applied at the lowest level, most nodes need to be updated, leading to only small gains in computational efficiency.%

\subsection{Network Details}
While the method described in Sec. \ref{sec:approach} would allow to sparsely update any kind of graph convolution, we found that spline convolutions \cite{Fey2018cvpr} find a balance between computational complexity and predictive accuracy. In contrast to the standard graph convolutions \cite{Kipf17iclr} used in \cite{Li21iccv}, spline convolutions maintain spatial information in the encoding by using a B-spline-based kernel function in the positional vertex space. This means that spline convolutions also take the relative position of neighboring nodes into account, a feature which is ignored in standard GNN-based methods like\cite{Li21iccv}. We use voxel-grid-based max-pooling \cite{Simonovsky17cvpr} due to its computational efficiency and simplicity. The method in \cite{Simonovsky17cvpr} clusters the graph's vertices by mapping them to a uniformly spaced, spatio-temporal voxel grid, with all vertices in a voxel being assigned to one cluster. \rev{In this work we use voxels of size $12\times 16\times 16$}. For each voxel, a node is sampled, resulting in the nodes of the coarse graph. Evaluating the effect of the clustering method on the overall network performance remains open for future work. Furthermore, we sub-sample the input event stream using uniform sampling to a fixed number of events. We found that other, more sophisticated sampling methods, such as non-uniform grid sampling \cite{Bi19iccv}, only marginally improved the performance, while being much more costly to compute.

Our model architecture is shown in Figure \ref{fig:pipeline}. It consists of $7$ graph convolution blocks (see Figure \ref{fig:pipeline}, bottom right) and $2$ pooling layers. For detailed information about our model architecture, we refer to the supplementary material.

\begin{figure*}[]
    \centering
    \begin{tabular}{ccc}
        \includegraphics[height=0.19\textwidth]{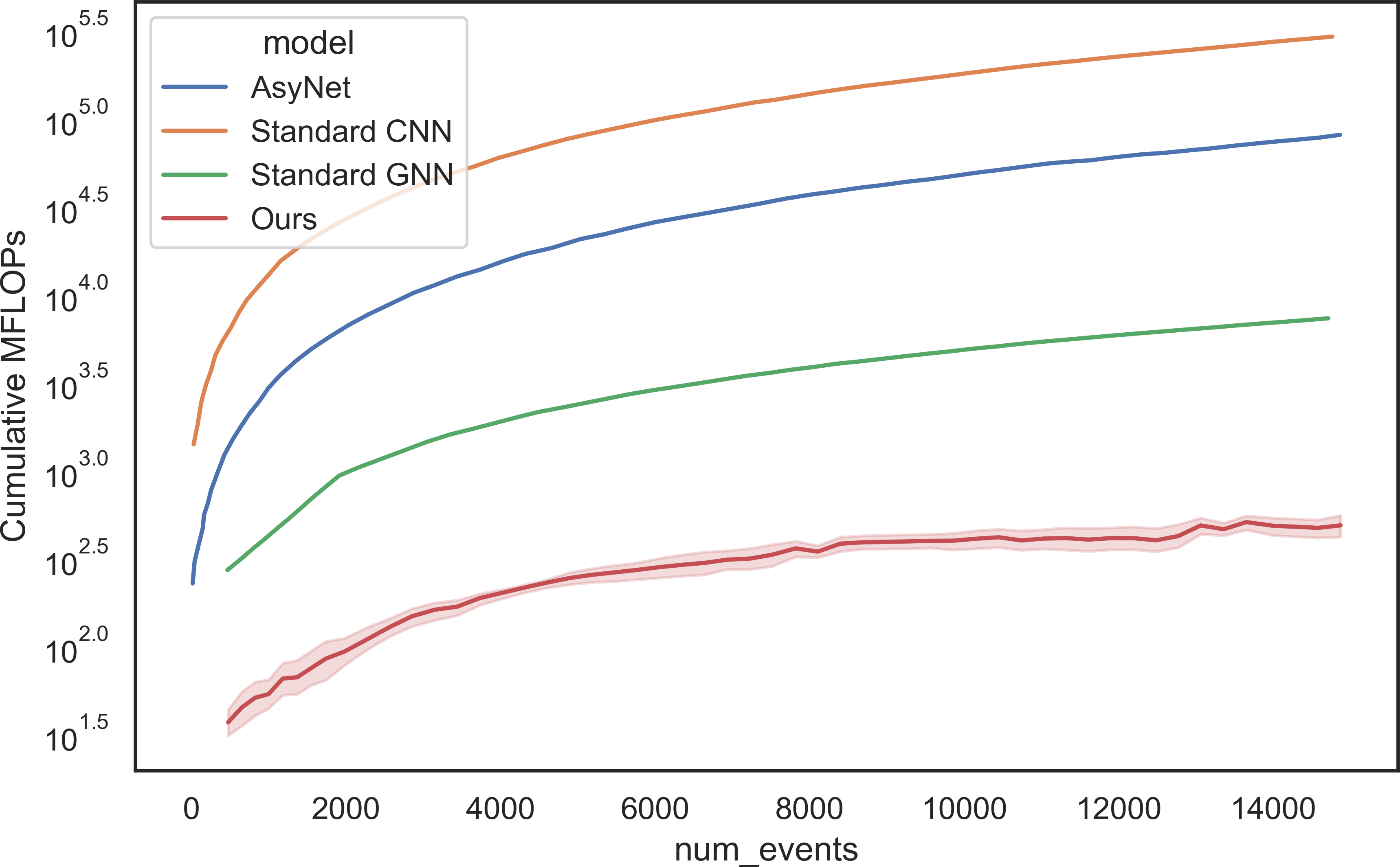}&
        \includegraphics[height=0.19\textwidth]{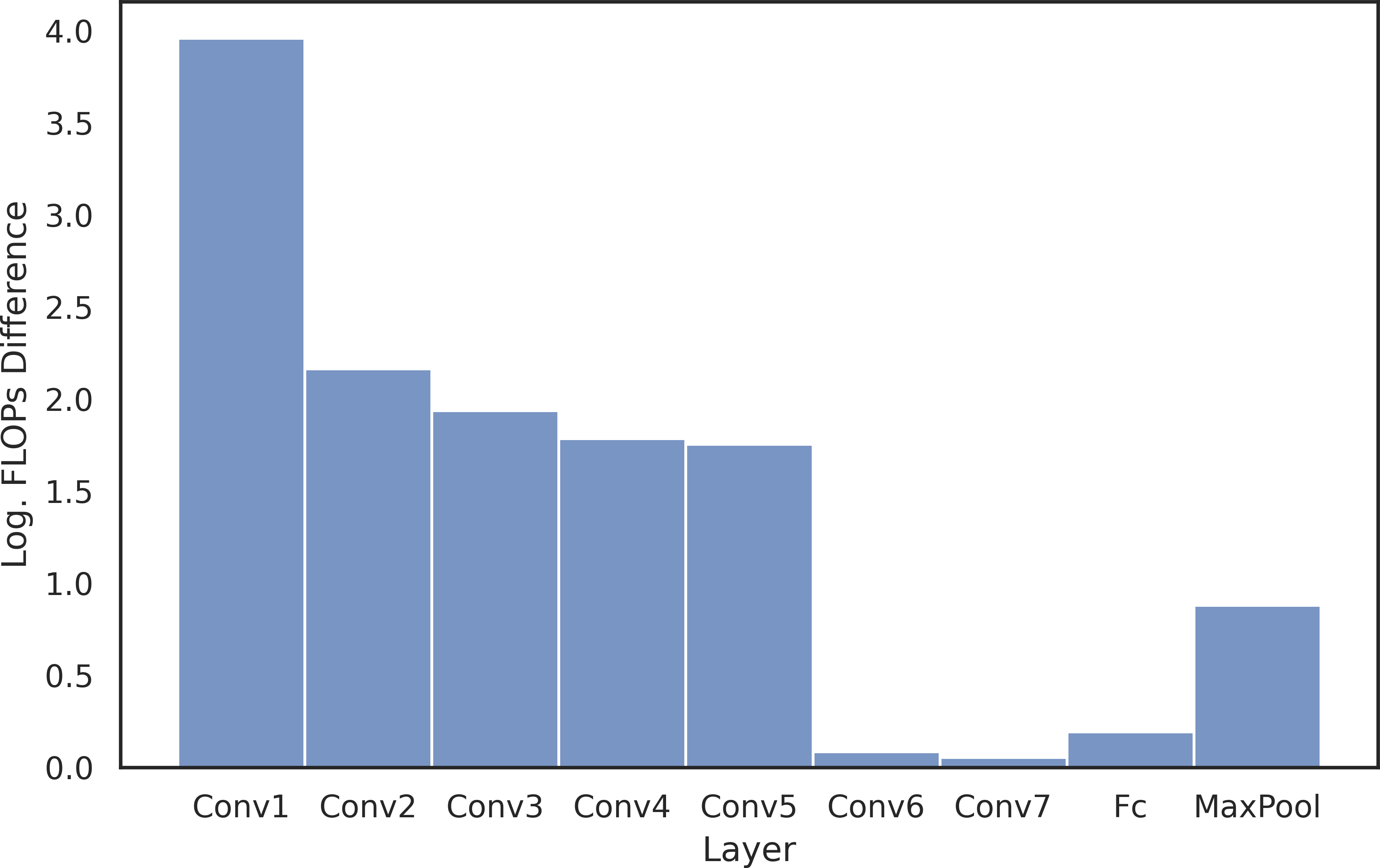}&
        \includegraphics[height=0.19\textwidth]{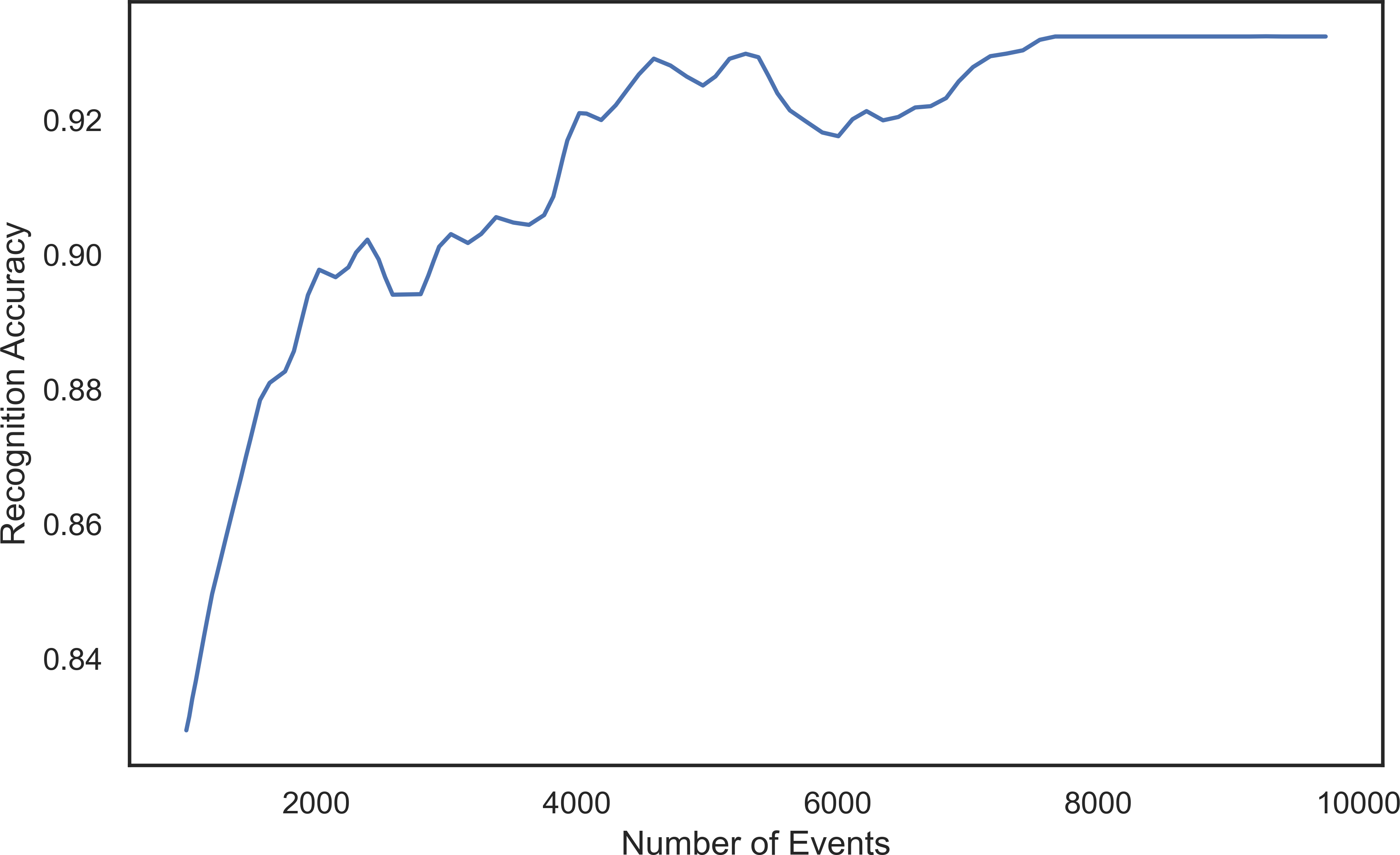}\\
        (a) MFLOPS over events& (b) MFLOP savings per layer&(c) Accuracy over events 
    \end{tabular}
    \caption{Computational savings of our method compared to a dense CNN, GNN and the method in \cite{Messikommer20eccv} on N-Cars\cite{Sironi18cvpr}.
    We compare the cumulative FLOPS for processing events in sequence (a). Here it is visible that already using a GNN reduces the number of FLOPS by a factor of 10. By additionally using our asynchronous formulation, we further reduce this number by a factor of 30. Additionally, for our method, computation grows much more slowly than for other methods.
    We show in (b) the FLOPS saved per layer, compared to a dense GNN. We see that our method saves most of the computation in the early and middle layers, where high feature dimensions are used. 
    Finally, we demonstrate the use of our method for early prediction (c).
    Although the model was trained with $10,000$ events, merely $2,500$ events are required to achieve over $90 \%$ accuracy.}
    \label{fig:computation}
    \vspace{-2ex}
\end{figure*}

\section{Experiments}
\label{sec:experiments}
All experiments within this work have been conducted using the PyG library \cite{Fey2019iclr} in the Torch framework \cite{Paszke17nipsw}. For training, we use the Lightning framework \cite{Falcon19software}.\

\textbf{Implementation Details:}
We used Adam \cite{Kingma15iclr} with batch size $16$ and an initial learning rate $10^{-3}$, which decreases by a factor of $10$ after $20$ epochs. We apply AEGNN to the tasks of object recognition and object detection.

We have analytically deduced the computational complexity of a forward pass of our model by adding up the computational complexity of each layer. A detailed derivation can be found in the supplementary material.

\subsection{Object Recognition}
\label{sec:experiments/recognition}
Event-based object recognition tackles the problem of predicting an object category from the event stream and is an important application of event cameras. Due to their high dynamic range and high temporal resolution, event cameras have the potential to detect objects, that would otherwise be undetectable by frame-based methods, especially in low-light conditions, or in conditions with severe motion blur.
We demonstrate that our approach is capable of solving this task very efficiently while achieving state-of-the-art recognition performance. The model is evaluated on two diverse datasets: The Neuromorphic N-Caltech101 dataset \cite{Orchard15fns} contains event streams recorded with a real event camera representing $101$ object categories in $8,246$ event sequences. each $300$ ms long, mirroring the well-known Caltech101 dataset\cite{FeiFei04cvprw} for images.  
The N-Cars dataset \cite{Sironi18cvpr} has real events, assigned to either a car or the background. It has $24,029$ event sequences, each being $100$ ms long. For training, we use the cross-entropy loss with batch-size $64$ (N-Cars) and $16$ (N-Caltech101).

\textbf{Recognition Performance} We compare AEGNN against several state-of-the-art methods, both asynchronous and synchronous, with different event representations (Tab. \ref{tab:performance_recognition}). 
We term methods as synchronous, if they require recomputation at each new event, and asynchronous otherwise. 
For quantitative comparison, we state the recognition accuracy on the test set. 
To assess the computational efficiency of each method, we process windows with $25,000$ events and measure the floating-point operations (FLOPs) required to update the prediction for each additional event. \\
H-First \cite{Orchard15pami}, HOTS \cite{Lagorce17pami}, HATS \cite{Sironi18cvpr} and DART \cite{Ramesh17arxiv} propose hand-designed features for object recognition. Typically, they are computationally efficient, but widely outperformed by our data-driven method. EST \cite{Gehrig19iccv} is a learnable and dense event representation that is jointly optimized with the downstream task. Although yielding very good recognition accuracy, it introduces additional data processing by using a learned representation and cannot be formulated asynchronously. Thus,  our method is $3,000$ times more efficient while achieving a similar predictive performance on N-Cars. AsyNet \cite{Messikommer20eccv} proposes an asynchronous, sparse network based on event-histograms. Hence, it does not explicitly account for the event's temporal component. Lastly, NVS-S and EvS-B \cite{Li21iccv} also use a graph-based event representation. In contrast to the standard graph convolutions used in EvS-B, the spline convolutions AEGNN encode spatial information. Consequently, our method is $21$ times more efficient while achieving a similar accuracy, in comparison to \cite{Li21iccv}.

\textbf{Scalability} While previously assuming a constant number of input events, in the following, we analyze the impact the number of events has on both the computational complex and the recognition accuracy to determine the viability of our method for low-latency prediction. 
To do this, we compare our model's test set accuracy on N-Cars for different numbers of events, and plot the accuracy and required cumulative computation in Figs. \ref{fig:computation} (a) and  (c). 
To highlight the efficiency of our method, we also plot the required number of FLOPs for the dense GNN, the asynchronous method \cite{Messikommer20eccv} and its dense, synchronous variant.
Our proposed method outperforms \cite{Messikommer20eccv} in terms of accuracy (Tab.\ref{tab:performance_recognition}) and in terms of FLOPs (Fig. \ref{fig:computation} (a)), showing a computation reduction by a factor of 300. 
The computational savings come from the comparably flat architecture and sparse graph representation. %
Notably, our model does not require the full event stream, that it was trained on, for a correct prediction. As demonstrated in Fig \ref{fig:computation} (c), only $5,000$ events are required to achieve state-of-the-art recognition accuracy, further improving the computational efficiency of our method. \rev{Moreover, our method takes $30\pm 4.8$ kFLOPS/ev for 25'000 events, averaged over all sequences. The low variance indicates a high level of stability.}

\begin{table}[H]
\centering
\resizebox*{\linewidth}{!}{
    \begin{tabular}{lccccc}
    \textbf{Model} & $2000$ & $4000$ & $6000$ & $8000$ \\
    \hline
    GNN & $1.72\cdot 10^3$ & $4.68\cdot 10^3$ & $7.89\cdot 10^3$ & $11.2\cdot 10^3$ \\
    \textbf{AEGNN (ours)} & \textbf{4.29} & \textbf{5.56} & \textbf{5.94} & \textbf{6.11}
    \end{tabular}}
\caption{Computational effort in MFLOPs per event of our sparse method compared to its dense equivalent, evaluated on NCaltech101. With a higher number of events, and thus increasing complexity of the event graph, the computational gap becomes larger.}
\label{fig:gnn_vs_aegnn}
\end{table}

\begin{table*}[t]
    \centering
    \begin{tabular}{lccccccc}
    & & & \multicolumn{2}{c}{N-Caltech101} & & \multicolumn{2}{c}{Gen1} \\
    \cline{4-5} \cline{7-8}
    Methods & Representation & Async. & mAP $\uparrow$ & MFLOP/ev $\downarrow$ & &  mAP $\uparrow$ & MFLOP/ev $\downarrow$ \\
    \hline
    YOLE \cite{Cannici19cvprw} & Event-Histogram & \cmark & 0.398 & 3682 & & - & - \\
    Asynet \cite{Messikommer20eccv} & Event-Histogram & \cmark &\textbf{0.643} & 200 & & 0.129 & 205 \\
    RED \cite{Perot20nips} & Event-Volume & \xmark & - & - & & \textbf{0.40} & 4712 \\
    NVS-S \cite{Li21iccv} & Graph & \cmark & 0.346$^*$ & 7.8 & & 0.086$^*$ & 7.8 \\
    \hline
    \textbf{Ours} & Graph & \cmark & 0.595 & \textbf{7.41} & & 0.163 & \textbf{5.26}
    \end{tabular}
    \caption{Comparison with several asynchronous and dense methods for object detection. The method in \cite{Li21iccv} was re-implemented and trained by us, as \cite{Li21iccv} only reports results for the object recognition task.}
    \label{tab:performance_detection}
\end{table*}
\subsection{Object Detection}
\label{sec:experiments/detection}
Event-based object detection seeks to classify and detect object bounding boxes from an event stream and is an emerging topic in event-based vision. Especially in night-time scenarios or when objects travel at high speeds, frame-based object detection degrades due to image degradation, caused by underexposure or severe motion blur. Event cameras by contrast do not suffer from these issues and are thus viable alternatives in these cases. We apply our framework to this task and validate our approach on two challenging datasets: the N-Caltech101 dataset \cite{Orchard15fns}, see Sec. \ref{sec:experiments/recognition}, and the Gen1 dataset \cite{Tournemire20arxiv}. While N-Caltech101 contains only one bounding box per sample, it contains 101 classes, making it a difficult classification task. By contrast, Gen1 targets an automotive scenario in an urban environment with annotated pedestrians and cars. With $228,123$ bounding boxes for cars and $27,658$ for pedestrians, the Gen1 dataset is much larger. To avoid the well-known over-smoothing problem of GNNs \cite{Li18aaai}, we adopt the same backbone as for the recognition task but use a YOLO-based object detection head \cite{Redmon16cvpr}, as illustrated in Fig. \ref{fig:pipeline}. Similar to \cite{Redmon16cvpr} we use a weighted sum of class, bounding box offset and shape as well as prediction confidence losses. 

\begin{figure*}[t]
\vspace{-1ex}
\centering
\begin{tabular}{c}
\includegraphics[width=0.3\textwidth]{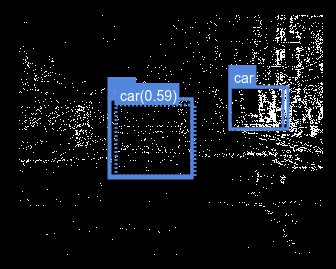}
\includegraphics[width=0.3\textwidth]{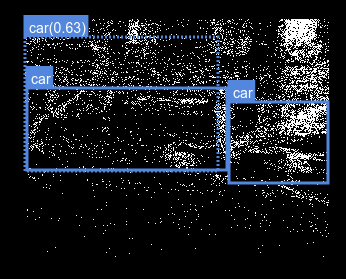}
\includegraphics[width=0.3\textwidth, height=4.2cm]{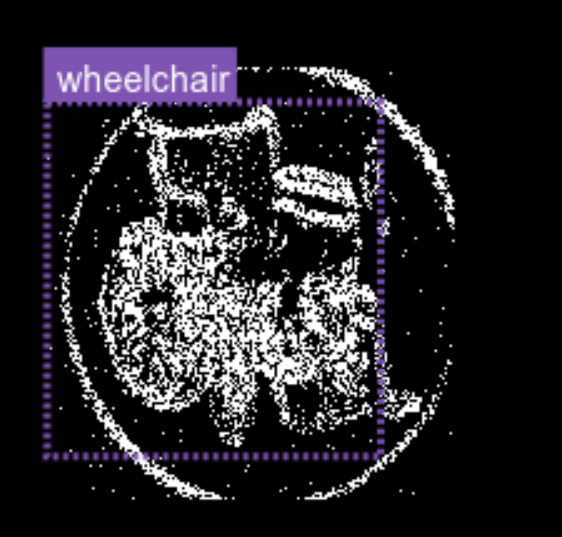}
\end{tabular}
\caption{Qualitative results of the object detection performed by our model on Gen1\cite{Tournemire20arxiv} and N-Caltech101\cite{Orchard15fns} dataset. Our predictions are shown as a dashed line, the labels as solid line. }
\label{fig:detection_example}
\vspace{-2ex}
\end{figure*}

\textbf{Detection Performance} To evaluate the performance of our model, we use the eleven-point mean average precision (mAP) \cite{Lin14eccv} score as  well as the computational complexity per event, as described in Sec. \ref{sec:experiments/recognition}. We compare with synchronous and asynchronous state-of-the-art methods and present the results in Tab.\ref{tab:performance_detection}. Qualitative results of our object detector on N-Caltech101 and the Gen1 dataset are shown in Fig.\ref{fig:detection_example}  We reimplement NVS-S \cite{Li21iccv}, as open-source code is not available.

Our method outperforms NVS-S \cite{Li21iccv} by 7.7\%, while using 21 times less computation. This is because NVS-S uses standard graph convolutions, and thus have a receptive field that is limited to their direct neighborhood, which deteriorates detection performance. Compared to RED \cite{Perot20nips}, we achieve a lower accuracy but outperform the method by a significant margin: While our method uses 0.39 MFLOPs/ev, \cite{Perot20nips} uses $4712$ MFLOPs/ev. This is because \cite{Perot20nips} uses a dense, synchronous recurrent network, and it is thus not capable of event-by-event processing. Finally, AsyNet\cite{Messikommer20eccv} outperforms AEGNN on N-Caltech101 by 4.8 mAP, but we show a 3.4 mAP higher performance on Gen1. While performances are comparable, we achieve this with  520-540 times fewer MFLOPs per event.

\textbf{Timing Experiments} We timed our method, implemented in Python and CUDA, on an Nvidia Quadro RTX. \rev{To construct the graph we implemented the radius search algorithm in \cite{Li21iccv} in CUDA, which takes 2 ms to generate a graph with 2,500 nodes.} For processing one event in an event graph of $4,000$ from N-Caltech101, the dense update requires $167 ms$, our sparse method $92 ms$. For $25,000$ events, the dense GNN needs $1014 ms$, our sparse method $129 ms$, an improvement by a factor of 8. A dense CNN with the same input requires $202 ms$. While our method is only 1.5 times faster than a CNN, we point out here that CNNs have highly optimized implementations in the PyTorch Library\cite{Paszke17nipsw}. However, we expect that if implemented on suitable hardware, such as FPGA or IPU\cite{Jia2019DissectingTG} processors, 
the reported computation reduction will lead to significant reductions in latency and power consumption, as was already demonstrated in \cite{Chang20jestcs}.

\section{Conclusion}
\label{sec:conclusion}
While event-based vision has made significant strides by adopting standard learning-based methods based on CNNs, these discard the spatio-temporal sparsity of events, which leads to wasteful computation. For this reason, geometric-learning approaches for event-based vision have gained in popularity. In this work, we introduced AEGNNs, which model events as evolving spatio-temporal graphs and formulate efficient update rules for each new event that restrict recomputation of network activations only to a few nodes, which are propagated to lower layers. We applied AEGNNs to the tasks of object recognition and detection. While in object recognition we achieved an up to a 11-fold reduction in computational complexity (FLOPs), for object detection we achieved an up to 32\% reduction, while outperforming asynchronous methods by 3.4\% mAP. We showed that this computation reduction speeds up processing latency by a factor of 8 compared to dense GNNs. 
We believe that, if our method is implemented on specialized hardware such as FPGA or IPUs\cite{Jia2019DissectingTG}, we will see additional reductions in latency and a significant reduction in power consumption.

\section{Acknowledgment}
This work was supported Huawei, and as a part of NCCR Robotics, a National Centre of Competence in Research, funded by the Swiss National Science Foundation (grant number 51NF40\_185543).

\section{Appendix}
Here we report additional information to support the main manuscript. In what follows, 
we will refer to figures, tables, sections, and equations from the manuscript by prepending "M-". We start by providing a detailed network overview in Sec.\ref{sec:app:network_details}. We then show a derivation for the number of FLOPS required to compute the Spline Convolution in \cite{Fey2018cvpr} for a single node in Sec.\ref{sec:app:comp_complexity}.
We finally list the licenses of the datasets used in this submission in Sec. \ref{sec:app:licenses}.

\subsection{Network Details}
\label{sec:app:network_details}

We use two network architectures in this work, one for object recognition (Sec. M-5.2) and one for object detection (Sec. M-5.3). Both networks consist of convolutional blocks, each containing a SplineConv \cite{Fey2018cvpr}, defined by the number of output channels $M_{out}$ and kernel size $k$, an ELU activation function, and a batch norm. As shown in Figure M-2, max graph pooling layers after the fifth and seventh convolution, as well as skip connections after the fourth and fifth convolution, are used. Also, a fully connected layer maps the extracted feature maps to the network outputs. 
The recognition network has convolutions with kernel size $k = 2$ and output channels $M_{out}^i = (1, 8, 16, 16, 16, 32, 32, 32).$ The convolutions in the detection network have a much larger kernel size $k = 8$ and more output channels $M_{out}^i = (1, 16, 32, 32, 32, 128, 128, 128).$ 

\subsection{Spline Convolutions Complexity}
\label{sec:app:comp_complexity}
In this work, we make heavy use of Spline Convolutions\cite{Fey2018cvpr}. 
Compared to standard GNN layers which only aggregate features over layers, Spline Convolutions also take into account the spatial arrangement of these neighbors and thus produce richer features. Here we will give a summary of spline convolutions and refer the reader to \cite{Fey2018cvpr} for more details.
Given nodes $i$ with features $\textbf{f}(i)$ we define convolution kernels $g_n$, with $n=1,...,M_{\text{out}}$ the index of the output feature. They act as
\begin{equation}
\label{eq:app:spline_conv}
(g_n*f)(i)=\frac{1}{\vert N(i)\vert}\sum_{l=1}^{M_{in}}\sum_{j\in N(i)}f_l(j)g_{n,l}(u(i,j))
\end{equation}

Here $f_l(j)$ is the input feature with index $j$, $N(i)$ counts the number of neighbors of node $i$,  and $u(i,j)$ are pseudo coordinates. These are defined as the normalized distance vector between nodes $i$ and $j$. 

The function $g$ is expanded as 
\begin{equation}
g_{n,l}(u) = \sum_{p\in P}w_{p,l,n}B^m_{p}(u)
\end{equation}
Here $P$ denotes an index set, which is a regular grid in 3 dimensions. It has two elements in each direction, resulting in $2^3=8$ elements (tuples). For each coordinate tuple a learnable weight $w_{p,l,n}$ is stored and multiplied by a B-Spline basis $B^m_{p}(u)$ in three dimensions. Each B-Spline basis is computed by forming the product of three splines as
\begin{equation}
B^m_p(u)=\prod_{s=1}^3 N^m_{p_s,s}(u_s),
\end{equation}
\emph{i.e.} one for each dimension. Here $m$ is the degree of the B-Spline, and in this work we use $m=3$.
Each function $N^m_{p_s,s}(u_s)$ can thus be written $N^m_{p_s,s}(u_s)=\sum^{m-1}_{j=0}a_ju_s^j$. 

\subsubsection{FLOPS Computation}
Here we count the FLOPS necessary. In what follows we define $N_i=\vert N(i)\vert$ and $N_p=\vert P\vert$ and will proceed in a series of steps. 
To evaluate Eq. \ref{eq:app:spline_conv} we first compute $u$ for all neighbors and dimensions, resulting in:
\begin{equation}
    FLOPS(u(i,j)) = N_i d
\end{equation}
then we compute the FLOPS to evaluate $B_p^m(u)$ as 
\begin{equation}
    FLOPS(B_p^m(u)) = 2mN_id + N_i (d-1)
\end{equation}
For the first term we make use of Horner's method\cite{Horner19PTRSL}, which states the optimal number of additions and multiplications for a polynomial of degree $m$ as $2m$. For the second term we count the FLOPS to compute the product. Each operation needs to be repeated for each neighbor.
Next we compute $g_{n,l}(u)$ as the sum of products over elements of $P$, input features and output features. 
\begin{equation}
    FLOPS(g_{n,l}(u)) = (2N_p-1)M_\text{out}M_\text{in}N_i
\end{equation}
Finally we aggregate these terms, first over neighbors and then over input features
\begin{align*}
    FLOPS\left(\sum_{l=1}^{M_\text{in}}\sum_{j\in N(i)}f_l(j)g_{n,l}(u(i,j)\right) =\\ (2N_i-1)M_\text{out}M_\text{in} + (M_\text{in}-1)M_\text{out} 
\end{align*}

Where the first term counts products and summation over neighbors, and the second counts summation over input features.
Finally, we divide all output features by $N_i$, adding additional $M_\text{out}$ FLOPS. We thus have 
\begin{align*}
    C_\text{tot}=&N_i d +2mN_i d +N_i (d-1)\\ &+(2N_p-1)M_{out}M_{in}N_i\\
    &+(2N_i-1)M_{out}M_{in}\\
    &+(M_{in}-1)M_{out} +M_{out}\\
=&N_i M_{out}M_{in}(1+2N_p) + N_i (2d+2md-1).
\end{align*}

\subsection{Licenses}
\label{sec:app:licenses}
We use the Prophesee Gen1 dataset\cite{Tournemire20arxiv} under the ``PROPHESEE GEN1 AUTOMOTIVE DETECTION DATASET LICENSE TERMS AND CONDITIONS" found at the URL \url{https://www.prophesee.ai/2020/01/24/prophesee-gen1-automotive-detection-dataset/}.
N-Caltech101\cite{Orchard15fns} is used under the ``Creative Commons Attribution 4.0 license" and downloaded at \url{https://www.garrickorchard.com/datasets/n-caltech101}. Finally, the N-Cars dataset\cite{Sironi18cvpr} is also released by Prophesee under \url{https://www.prophesee.ai/2018/03/13/dataset-n-cars/}.

{\small
\bibliographystyle{ieee_fullname}
\bibliography{paper}
}

\end{document}